\documentclass[a4paper]{article}

\usepackage[english]{babel}
\usepackage[utf8x]{inputenc}
\usepackage[T1]{fontenc}

\usepackage[a4paper,top=3cm,bottom=2cm,left=3cm,right=3cm,marginparwidth=1.75cm]{geometry}

\usepackage{amsmath}
\usepackage{graphicx}
\usepackage[colorinlistoftodos]{todonotes}
\usepackage[colorlinks=true, allcolors=blue]{hyperref}

\title{An Automatic Diagnosis Method of Facial Acne Vulgaris Based on Convolutional Neural Network}

\author{Xiaolei Shen, Jiachi Zhang, Chenjun Yan, Hong Zhou\thanks{Correspondence and requests for materials should be addressed to H.Z. (email:zhouh@mail.bme.zju.edu.cn) }\\
Key Laboratory for Biomedical Engineering of Ministry of Education, \\
Zhejiang University of China }

\begin{document}
\maketitle

\begin{abstract}
In this paper, we present a new automatic diagnosis method of facial acne vulgaris based on convolutional neural network. This method is proposed to overcome the shortcoming of classification types in previous methods. The core of our method is to extract features of images based on convolutional neural network and achieve classification by classifier. We design a binary classifier of skin-and-non-skin to detect skin area and a seven-classifier to achieve the classification of facial acne vulgaris and healthy skin. In the experiment, we compared the effectiveness of our convolutional neural network and the pre-trained VGG16 neural network on the ImageNet dataset. And we use the ROC curve and normal confusion matrix to evaluate the performance of the binary classifier and the seven-classifier. The results of our experiment show that the pre-trained VGG16 neural network is more effective in extracting image features. The classifiers based on the pre-trained VGG16 neural network achieve the skin detection and acne classification and have good robustness.

\end{abstract}

\section{Introduction}

Acne is a chronic skin disease whose symptoms and disease mechanisms vary much. Because of its diversity, we cannot find a unified classification method. In this article, we adopt a type of classification consisting of non-inflammatory lesions (open comedones which also can be called blackheads and closed comedones which is also named whiteheads), inflammatory lesions (papules and pustules) and relatively more severe types, nodules and cysts\cite{1}. The data sets and the classifier in this article are all designed to detect these six types of acne vulgaris (blackheads, whiteheads, papules, pustules, modules and cysts) mentioned above on human faces. A patient’s face usually suffers from multiple types of acne lesions at the same time. The causes and treatments of different acne lesions are different, so it is necessary to make an accurate and objective diagnosis of the face with acne vulgaris before treatment\cite{2}. The manual observation and count of acne is the traditional diagnosis method. However, the method is not an effective method. It is labor-intensive, time-consuming and subjective because the diagnosis results depend on expert’s experience and ability. Therefore, a computerized method is very necessary. Researchers have done much research on the automatic diagnosis of facial acne with image processing techniques and machine learning theories.

Automatic diagnosis of facial acne needs to achieve the detection and classification of the region of interest(ROI). The detection of ROI mainly refers to the detection of facial skin. Previous automatic diagnosis methods also include specific positioning of various acne. Common skin detection models are based on special color spaces to achieve, such as RGB, HSV, YCbCr\cite{3} \cite{4} \cite{5} \cite{6} \cite{7}. Due to mixing of chrominance and luminance data, RGB is not a good choice for skin detection. Although YCbCr avoids this problem, its actual detection effect is still unstable and susceptible to some environmental influences\cite{8}. In past automatic diagnosis methods, specific positioning of the acne area is necessary. Chantharaphaichit et al. proposed a facial acne area detection method based on grayscale and HSV color space\cite{9}. But color, shape and lighting conditions of acne have a great influence on the detection effect with this method. Kittigul and Uyyanonvara proposed a different method based on Heat-Mapping and adaptive thresholding\cite{10}. However, the detection results still contain some noise, such as the area of mouth and skin. The above methods are based on special color spaces and thresholds to achieve the detection of acne area. However, the defect that these methods are too dependent on the threshold segmentation and lack of universality is non-negligible.

Previous automatic diagnosis methods of acne are mostly based on feature extraction to achieve classification of acne. Chang and Liao, and Malik et al. proposed two similar approaches both based on support vector machine and feature extraction to achieve the classification\cite{11} \cite{12}. Furthermore, Malik et al. classified the severity levels into mild, moderate, severe and very severe. Because of the limitation of classification, these methods can’t achieve the integral analysis with a detailed description.

Convolutional neural network has been widely applied to image classification. Sometimes it presents a high recognition ability and gives a better performance in certain projects such as the recognition of traffic signs, faces and handwritten digits than human beings\cite{13} \cite{14} \cite{15}. In recent years, with a lot of research work in the ImageNet dataset, the image recognition based on CNN has been continuously improved. In recent studies, CNN with a deeper and wider structure has a large number of parameters. So, the neural network is easy to over fit in the training process. To prevent over fitting, Hinton et al. proposed the “dropout” method and the data augmentation method\cite{16} \cite{17} \cite{18}.

In this paper, we propose a novel automatic diagnosis method to overcome the shortcoming of classification categories in the previous methods and achieve a holistic analysis of facial acne vulgaris with a detailed description. Different from the old methods, our method extracts features by CNN instead of manual selection. With the novel method, more and deeper features are extracted to enhance classification accuracy and add more classification types\cite{19} \cite{20}. The major research work in this paper is to achieve the detection of facial skin and the seven-type-classification of facial acne vulgaris by the classifiers. In the experiment, we compare the validity of extracting features between the CNN model constructed manually by ourselves and the VGG16 model which has been pre-trained on ImageNet. We select the excellent performance model to achieve facial skin area detection and acne 7 classification (including 6 types of acne and normal skin). On the basis of the classification results, we finally realizes the automatic diagnosis and the integral analysis of patients’ faces.

\section{Proposed methodology}

The process of the automatic diagnosis method based on CNN is shown in Figure 1. There are two main steps: 1) skin detection to locate ROI and 2) seven-type-classification of facial acne vulgaris to achieve the automatic acne diagnosis. The skin detection method is achieved by a binary classifier differentiating skin from non-skin. The binary classifier can extract the features of input images and classify them into skin and non-skin to achieve the skin detection. The acne classification method is achieved by a seven-classifier of facial acne vulgaris. The seven-type-classifier can extract the features of input images and classify them into one of the seven categories. Then, based on the classification results, integral analysis with a detailed description including acne categories and their respective proportions will be generated.

\subsection{The data sets, data augmentation and data preprocessing}

We collected two data sets for the binary classifier and the seven-classifier. 70\% of the each data set is used for training and the rest is used for validation. The data set of the binary classifier consists of 50x50 images cut from the original image of 500x500, containing 3000 skin images and 3000 non-skin images. Skin images include normal skin and diseased skin. A portion of the data set for the training and validation of the binary classifier is shown in the figure 2. The data set of the seven-classifier consists of 50x50 images cut from the original image of 500x500 and images augmentation based on this, including 6516 blackhead images, 6511 whitehead images, 6479 papules images, 6502 pustules Images, 6526 cysts, 6479 nodule images, 3310 normal skin images. A portion of the data set for the training and verification of the seven-classifier is shown in the figure 3. Furthermore, we also prepared a test data set in order to evaluate the models we trained. The test data is shown with the test result in section experiments and results.

The original data of seven-classification is small, and it’s very easy to lead the neural network over fitting. We use the simplest and most commonly used method data augmentation to prevent over fitting\cite{17}. In this paper, we used the transformation method of rotation, shift, shear, scaling and horizontal flip to enlarge our data sets. Transformation can be represented by a transformation matrix. There is a transformation matrix M,

\begin{equation}
M = \begin{bmatrix} A & B & C\\ D & E & F\\ G & H & I \end{bmatrix}
\end{equation}
A and E control the scaling of the image, C and F control the shift of the image, B and D control the shear of the image. Detailedly, the transform matrix of rotation is

\begin{equation}
M = \begin{bmatrix} cos\theta & -sin\theta & -cos\theta\cdot\frac{h+1}{2}+sin\theta\cdot \frac{w+1}{2}+\frac{h+1}{2}\\ sin\theta & cos\theta & -sin\theta\cdot\frac{h+1}{2}-cos\theta\cdot \frac{w+1}{2}+\frac{w+1}{2}\\ 0 & 0 & 1 \end{bmatrix}
\end{equation}

In the matrix, h is the height of the picture, w is the width of the picture, $\theta$ is the angle of rotation.
The transform matrix of shift is

\begin{equation}
M = \begin{bmatrix} 1 & 0 & tx\\ 0 & 1 & ty\\ 0 & 0 & 1 \end{bmatrix}
\end{equation}

In the matrix, tx is the shift size of height, ty is the shift size of width.
The transform matrix of shear is

\begin{equation}
M = \begin{bmatrix} 1 & -sin(shear) & -zx\cdot \frac{h+1}{2}+\frac{h+1}{2}\\ 0 & cos(shear) & -\frac{h+1}{2}-cos(shear)\cdot \frac{w+1}{2}+\frac{w+1}{2}\\ 0 & 0 & 1 \end{bmatrix}
\end{equation}

In the matrix, shear is the transformation intensity of shear, h is the height of the picture, w is the width of the picture.
The transform matrix of zoom is

\begin{equation}
M = \begin{bmatrix} zx & 0 & -zx\cdot \frac{h+1}{2}+\frac{h+1}{2}\\ 0 & zy & -zy\cdot \frac{w+1}{2}-\frac{w+1}{2}\\ 0 & 0 & 1 \end{bmatrix}
\end{equation}

In the matrix, zx and zy control the zoom of the image.
Moreover, we set the random horizontal flip image. The examples of transformation are shown in the figure 4.

Every original image has been transformed several times, but retained its key features. Additionally, the raw pixels of the images are located between 0-255. They already have a certain standard ability. In the experiment, we used a normalization approach to scale the pixel data to a range of 0-1, which is beneficial for the gradient descent in the training process.

\subsection{Export the feature vector}

Convolutional neural networks have the ability to extract image features. Normally, a deeper CNN can extract more specific and complex features. But it cannot ensure better performance of the model if it involves too many convolutional layers. VGG16 in the transfer learning is more commonly used convolutional neural network model. A VGG16 model without top layers is shown in Table 1. In the ImageNet and other large data sets VGG16 has an imposing identifying ability, and its effectiveness of the feature extraction has been proved valid on large data sets\cite{21}. In this paper, we use the pre-trained VGG16 network to extract the characteristics of the two classification data sets and the seven classification data sets, and express them with 512-dimension feature vector. The advantage of using pre-trained models is obvious. We can not only extract effective features but also save training time. And the validity of extraction feature of pre-trained model has been validated. Relatively speaking, the training of a capable network for extracting effective features needs to consume more resources.

In this paper, we also construct a CNN. The network structure is shown in Table 3. The net is trained and tested in the data set built by ourselves and realize the image feature extraction alike VGG16 network.

\subsection{Skin detection (binary classifier) and acne classification (seven-classifier)}

he detection of skin area is based on the binary classifier. Relatively, to build a binary classifier with high accuracy is not a difficult task, because the random classification has an accuracy of 50\%. The feature vector of the image is extracted by the CNN model, and the classifier is used to classify and output the probability of skin or non-skin to achieve the skin detection. Similarly, the classification of acne is based on the feature extraction by the CNN and classification by the seven-classifier, and output the probability of each acne class. The structure of the binary classifier is shown in Table 2. Seven-classifier structure is similar to the binary classifier. The only difference is the output vector at the final dense layer is 7-dimension. 

After training the seven-type-classifiers can classify six types of acne and healthy skin. Combined with skin-and-non-skin binary classifier, this classifier can achieve a facial acne diagnosis. This paper introduces sliding window method to achieve automatic cropping and traversing of the input facial images then classifier of each small area. Statistics of all classification results complete the overall description of facial acne vulgaris.

In this paper, we adopt some techniques to lower the risk of over fitting. We use the “ReLU” activation function and the “dropout” trick to prevent over fitting\cite{17}. Similarly, the data augmentation and the data normalization have the same effect.

\section{Experiment and results}
\subsection{Evaluation method}

ROC represents the "Receiver Operating Characteristic" curve, usually used to assess the performance of the binary classifier. And AUC represents the Area Under ROC Curve, normally, the AUC is larger, the classifier’s performance is better. The straight line connected by point (0,0) and point (1,1) represents a random binary classifier\cite{22} \cite{23}. The Youden's index is a summation of the ROC curve, assessing the validity of the diagnostic marker, and selecting the optimal segmentation threshold\cite{23} \cite{24}. ROC curve is very appropriate to evaluate the performance of binary classifier, but less credible to evaluate multi-classifier. In this paper, we do not use the extended form of the ROC curve to evaluate the performance of seven-classifier, but use the normalized confusion matrix as the method to evaluate the performance seven classifier. Normalized confusion matrix is normalized from the confusion matrix. Confusion matrix comprises of the numbers of each categories into which the images of our testing data set are classified, including both correct numbers and incorrect numbers\cite{25}.

\subsection{Skin detection and performance of binary classifier}

The ROC curve of the binary classifier based on the pre-trained VGG16 neural network is shown in Figure 5. The detailed index analysis is shown in Table 4 and the results of skin detection are shown in Figure 6. The non-skin areas in the original image including hair, eyes, background, etc. are well detected and covered with black blocks, and the skin areas in the original image are also retained as much as possible. Misjudgments mainly occur when the sliding window contains the junction area of skin area and non-skin. This is the major source of error during detection.

\subsection{Acne classification and the performance of seven-classifier}

In this paper, the normalized confusion matrix is used to evaluate the performance of seven-classifier based on the pre-trained VGG16 model on the testing data set. The normalized confusion matrix is shown in Figure 7. The overall description of facial acne symptoms in this article is shown in Figure 8. The overall description of the image predicted by our model and the overall description of the experts' diagnoses are shown in Table 5.

\section{Discussion}

Compared with traditional skin detection methods based on special color space, the skin detection method in this paper based on skin-and-non-skin binary classifier is more effective and more robust. Specifically, we find the skin detection method in this paper has a strong anti-interference ability to light conditions and color difference. Therefore it is more adaptive to many different conditions for the skin detection. However, it can be seen that there are still some hair areas at junction areas of skin and non-skin that cannot be well detected. The split boundaries in the test images have a clear zigzag shape. It looks bad, but in this experiment, it’s a side effect of an adaptive method, 50 × 50 cutting. And it rarely affects the accuracy of the acne detection in this article. In general, our skin detection method has a better performance in macro aspect, but its micro performance is not satisfying enough.

The seven-type-classifier is the core of the article and is the necessary step to achieve the overall description of facial acne vulgaris. The seven-type-classifier has managed to cut and classify skin areas of the input images. Its practical value is in a way proved through the performance indicators in the Table 5 and Figure 7. Compared with past methods, we break the restrictions of classification. We achieve the seven classification of facial acne vulgaris, realize the automatic diagnosis and give the integral analysis of patients’ face. The most important advancement is the number of the classification categories increased to seven, and propose a more detailed holistic analysis.

Generally speaking, if the acne traits are more obvious, the seven-classifier recognition ability is also stronger on this category. Because this type of acne can be better expressed by the 512-dimensional feature vector. In this article, the detection results of the healthy skin, papules, blackheads and whiteheads in the automatic diagnosis are better, and the detection results of cysts, pustules and nodules are relatively poor. The main problem is the 50x50 box cannot contain the whole information of cysts, pustules and nodules of which the disease area is large. The loss of information leads to the effect decline in detection results.

In this paper, we also make comparison of different models. We find the performance of extracting image features by the network we constructed and trained is weaker than the pre-trained VGG16 model. In the experiment, the ROC curve was used to evaluate the performance of the two classifiers based on two different CNN models. The experimental results show that the CNN we constructed and trained gains higher accuracy, but the detection result is poor in the actual test. We consider that the small data set and the high similarity of the augmentation data are the main reason. Considering that the data set in this paper is small, the generalized performance of CNN trained with small data sets is poor. So, we use the VGG16 CNN to successfully achieve the transfer learning on small data sets, and to complete the extraction of effective features in the images. Similarly, this situation occurs on seven-type-classifier.

In future research, we are prepared to build a more standardized data set, which is supposed to include more types of symptoms and conditions, to meet higher training and testing requirements of CNN and improve the classifiers’ generalization performance. As the data set grows stronger, we will consider to add more detailed description to the automatic diagnosis such as the severe level or the ages of acne vulgaris. In addition, we will consider the use of multiple pre-trained CNN model to replace the current single VGG16 neural network model, to make full use of each neural network model so that we can extract more effective features from the input images.

\bibliographystyle{unsrt}

\begin{thebibliography}{10}

\bibitem{1}
H.~C. Williams, R.~P. Dellavalle, and S~Garner.
\newblock Acne vulgaris.
\newblock {\em Lancet}, 379(9813):361--372, 2012.

\bibitem{2}
K~Bhate and H.~C. Williams.
\newblock Epidemiology of acne vulgaris.
\newblock {\em British Journal of Dermatology}, 168(3):474--85, 2013.

\bibitem{3}
R.~L. Hsu, M.~Abdelmottaleb, and A.~K. Jain.
\newblock Face detection in color images.
\newblock In {\em International Conference on Image Analysis and Recognition},
  pages 454--463, 2010.

\bibitem{4}
S.~L. Phung, A.~Bouzerdoum, and D.~Chai.
\newblock A novel skin color model in ycbcr color space and its application to
  human face detection.
\newblock In {\em International Conference on Image Processing. 2002.
  Proceedings}, pages I--289--I--292 vol.1, 2002.

\bibitem{5}
V.~Vezhnevets, V.~Sazonov, and A.~Andreeva.
\newblock A survey on pixel-based skin color detection techniques.
\newblock {\em IN PROC. GRAPHICON-2003}, pages 85--92, 2003.

\bibitem{6}
S.~L. Phung, A.~Bouzerdoum, and D.~Chai.
\newblock Skin segmentation using color pixel classification: analysis and
  comparison.
\newblock {\em IEEE Transactions on Pattern Analysis \& Machine Intelligence},
  27(1):148, 2005.

\bibitem{7}
M.~J. Jones and J.~M. Rehg.
\newblock Statistical color models with application to skin detection.
\newblock In {\em Computer Vision and Pattern Recognition, 1999. IEEE Computer
  Society Conference on}, page 1274, 1999.

\bibitem{8}
I.~Kim, H.~J. Shim, and J.~Yang.
\newblock Face detection, face detection project, ee368, 2003.

\bibitem{9}
T.~Chantharaphaichi, B.~Uyyanonvara, C.~Sinthanayothin, and A.~Nishihara.
\newblock Automatic acne detection for medical treatment.
\newblock In {\em Information and Communication Technology for Embedded
  Systems}, pages 1--6, 2015.

\bibitem{10}
N.~Kittigul and B.~Uyyanonvara.
\newblock Automatic acne detection system for medical treatment progress
  report.
\newblock In {\em Information and Communication Technology for Embedded
  Systems}, pages 41--44, 2016.

\bibitem{11}
C.~Y. Chang and H.~Y. Liao.
\newblock Automatic facial skin defects detection and recognition system.
\newblock In {\em Fifth International Conference on Genetic and Evolutionary
  Computing}, pages 260--263, 2011.

\bibitem{12}
A.~S. Malik, R.~Ramli, A.~F.~M. Hani, Y.~Salih, B.~B. Yap, and H.~Nisar.
\newblock Digital assessment of facial acne vulgaris.
\newblock In {\em Instrumentation and Measurement Technology Conference}, pages
  546--550, 2014.

\bibitem{13}
C.~Dan, U.~Meier, and J.~Schmidhuber.
\newblock Multi-column deep neural networks for image classification
  supplementary online material.
\newblock 157(10):3642--3649, 2012.

\bibitem{14}
Y.~Lecun, et~al.
\newblock Backpropagation applied to handwritten zip code recognition.
\newblock {\em Neural Computation}, 1(4):541--551, 1989.

\bibitem{15}
P.~Sermanet, S.~Chintala, and Y.~Lecun.
\newblock Convolutional neural networks applied to house numbers digit
  classification.
\newblock In {\em International Conference on Pattern Recognition}, pages
  3288--3291, 2013.

\bibitem{16}
G.~E. Hinton, N.~Srivastava, A.~Krizhevsky, I.~Sutskever, and R.~R.
  Salakhutdinov.
\newblock Improving neural networks by preventing co-adaptation of feature
  detectors.
\newblock {\em Computer Science}, 3(4):págs. 212--223, 2012.

\bibitem{17}
A.~Krizhevsky, I.~Sutskever, and G.~E. Hinton.
\newblock Imagenet classification with deep convolutional neural networks.
\newblock In {\em International Conference on Neural Information Processing
  Systems}, pages 1097--1105, 2012.

\bibitem{18}
K.~He, X.~Zhang, S.~Ren, and J.~Sun.
\newblock Delving deep into rectifiers: Surpassing human-level performance on
  imagenet classification.
\newblock pages 1026--1034, 2015.

\bibitem{19}
Y.~Lecun, K.~Kavukcuoglu, and C.~Farabet.
\newblock Convolutional networks and applications in vision.
\newblock In {\em IEEE International Symposium on Circuits and Systems}, pages
  253--256, 2010.

\bibitem{20}
Dan~C. C., U.~Meier, J.~Masci, L.~M. Gambardella, and Jürgen S.
\newblock High-performance neural networks for visual object classification.
\newblock {\em Computer Science}, 2011.

\bibitem{21}
K.~Simonyan and A.~Zisserman.
\newblock Very deep convolutional networks for large-scale image recognition.
\newblock {\em Computer Science}, 2014.

\bibitem{22}
T.~Fawcett.
\newblock An introduction to roc analysis.
\newblock {\em Pattern Recognition Letters}, 27(8):861--874, 2006.

\bibitem{23}
K.~Hajiantilaki.
\newblock Receiver operating characteristic (roc) curve analysis for medical
  diagnostic test evaluation.
\newblock {\em Caspian Journal of Internal Medicine}, 4(2):627, 2013.

\bibitem{24}
R.~Fluss, D.~Faraggi, and B.~Reiser.
\newblock Estimation of the youden index and its associated cutoff point.
\newblock {\em Biometrical Journal}, 47(4):458–472, 2005.

\bibitem{25}
F.~Pedregosa, et~al.
\newblock Scikit-learn: Machine learning in python.
\newblock {\em Journal of Machine Learning Research}, 12(10):2825--2830, 2013.

\end{thebibliography}

\textbf{Acknowledgements}\\
Thanks to the company Dr.Miao for the provision of data and the support of diagnostic knowledge. The datasets generated and analyzed during the current study are available from the corresponding author on reasonable request.

\textbf{Author Contributions Statement}\\
H.Z. conceived the experiments; X.L.S. and J.C.Z designed and performed the experiments; H.Z.,  X.L.S. and J.C.Z analyzed the data and results; C.J.Y, X.L.S and J.C.Z wrote the manuscript and prepared all tables and figures; These authors contributed equally to this work.

\textbf{Additional Information}
Competing financial interests: The authors declare no competing financial interests.

\newpage
\begin{figure}
\centering
\includegraphics[width=1.0\textwidth]{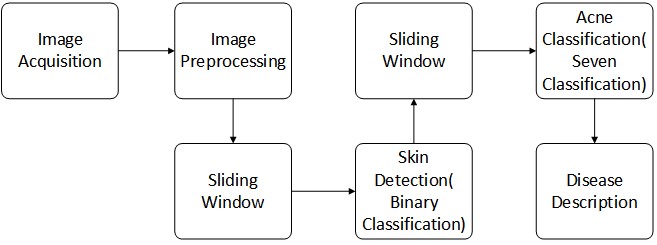}
\caption{\label{fig:process}The process of the automatic diagnosis method.(sec 2)}
\end{figure}

\begin{figure}
\centering
\includegraphics[width=1\textwidth]{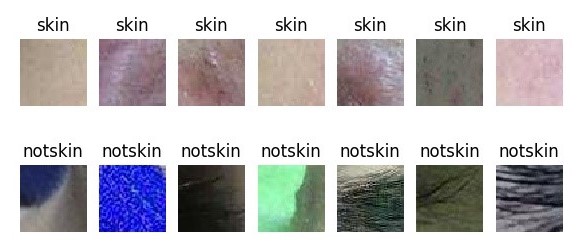}
\caption{\label{fig:skinandnotskin1}A part of the data set for the training and validation of the binary classifier.(sec 2.1)}
\end{figure}

\begin{figure}
\centering
\includegraphics[width=1\textwidth]{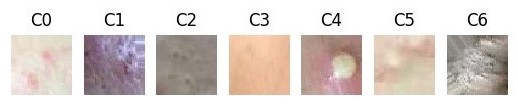}
\caption{\label{fig:seventypes1}A part of the data set for the training and verification of the seven-classifier.(sec 2.1) Papule (C0), cyst (C1), blackhead (C2), normal skin (C3), pustule (C4), whitehead (C5), nodule (C6)}
\end{figure}

\begin{figure}
\centering
\includegraphics[width=1\textwidth]{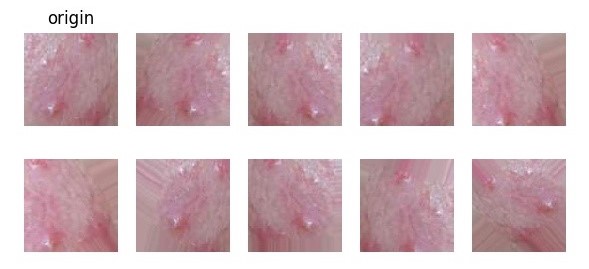}
\caption{\label{fig:transformation}The examples of transformation.(sec 2.1)}
\end{figure}

\begin{figure}
\centering
\includegraphics[width=1\textwidth]{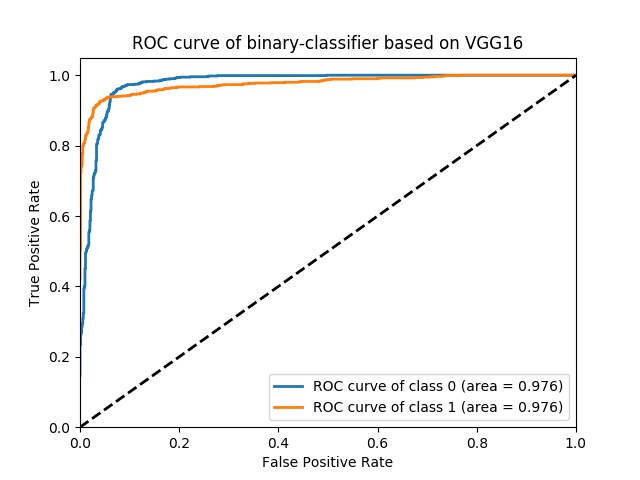}
\caption{\label{fig:ROC}The ROC of the skin-and-non-skin binary classifier.(sec 3.2) class 0 represents skin, class 1 represent non-skin. class 0 as positive, class 1 as negative (epochs=20). Y-axis is TPR (Sensitivity), X-axis is FPR (1-Specificity).}
\end{figure}

\begin{figure}
\centering
\includegraphics[width=1\textwidth]{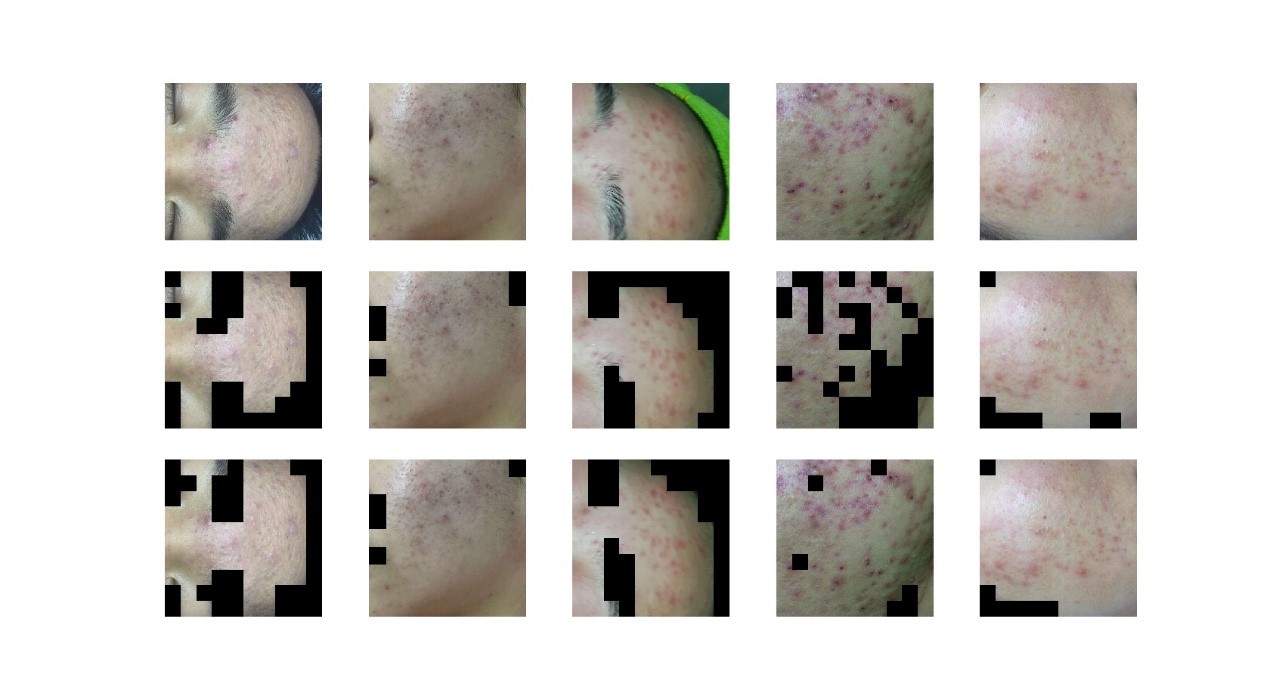}
\caption{\label{fig:skindetection}The results of skin detection by binary classifier.(sec 3.2) The first row presents five original images (500 x 500), each of which includes skin areas and non-skin areas. The second row presents five images covered by masks built through the binary classifier based on a manually constructed neural network. The mask is connected by 50 × 50 boxes. The third row presents five images covered by masks built through the binary classifier based on the VGG16 neural network. }
\end{figure}

\begin{figure}
\centering
\includegraphics[width=1\textwidth]{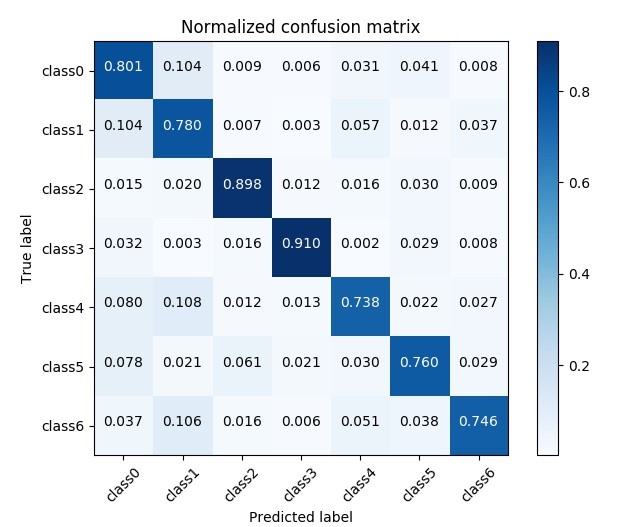}
\caption{\label{fig:confusionmatrix}The normal confusion matrix of the seven-classifier based on VGG16 model.(sec 3.3) Papule (class0), cyst (class1), blackhead (class2), normal skin (class3), pustule (class4), whitehead (class5), nodule (class6). Each column of the matrix represents the instances in a predicted class while each row represents the instances in an actual class.}
\end{figure}

\begin{figure}
\centering
\includegraphics[width=1\textwidth]{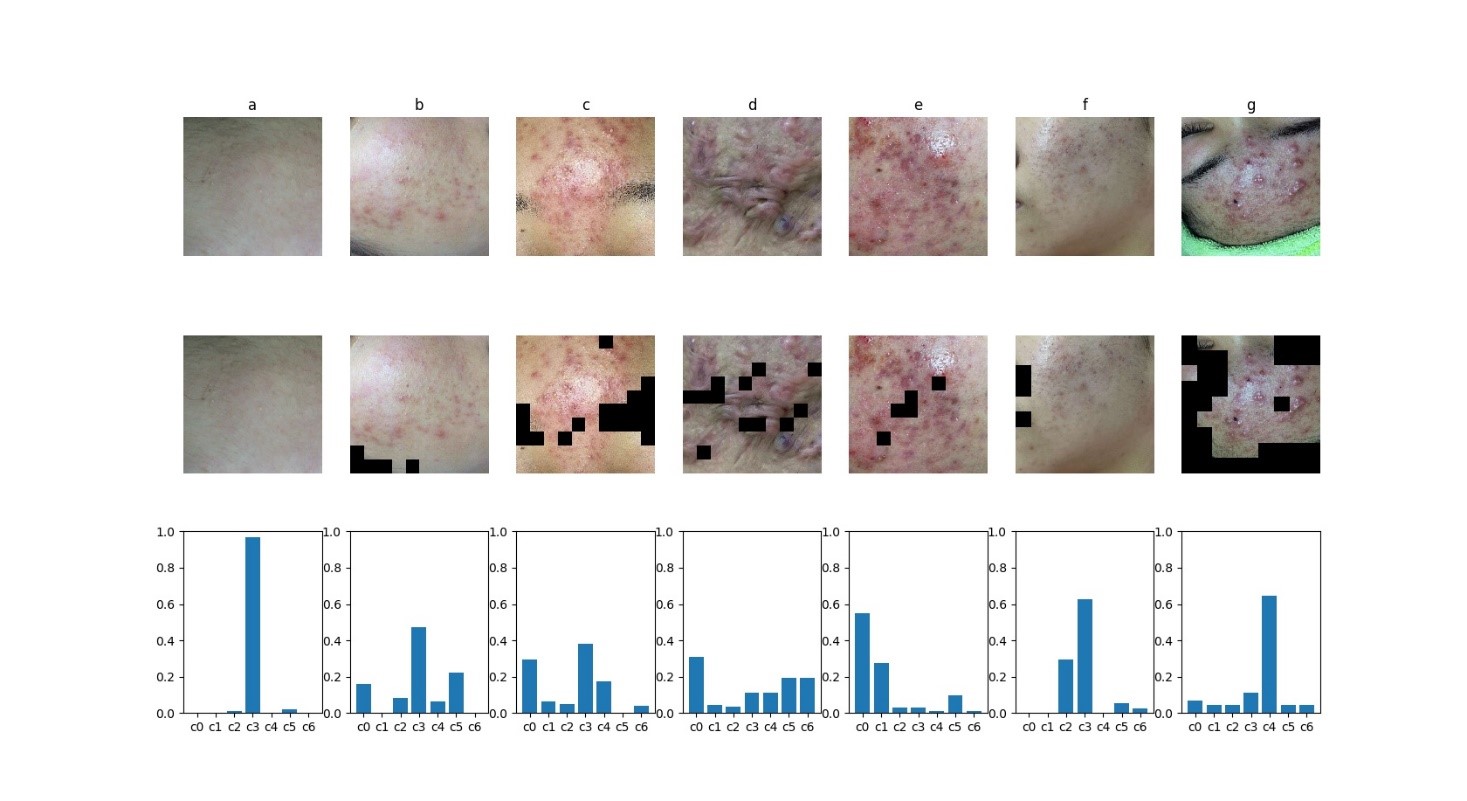}
\caption{\label{fig:sevenresult}The integral analysis of facial skin area.(sec 3.3) Seven images with masks are input images. The areas with black blocks are non-skin areas, the rest areas are skin areas. The result of integral analysis is shown with the histogram of corresponding proportions. Vertical coordinate of the histogram is the proportion, abscissa is the type of symptoms, followed by papule (c0), cyst (c1), blackhead (c2), normal skin (c3), pustule (c4), whitehead (c5), nodule (c6). }
\end{figure}

\newpage
\begin{table}[tbp]
\centering  
\begin{tabular}{ll}  
\hline
Layer & Type of layer \\ \hline  

Input (50$\times$50 RGB image)\\

Block1Conv1 & Conv2D-64\\
Block1Conv2	& Conv2D-64\\
BlockPool &	Maxpooling2D\\
Block2Conv1 &	Conv2D-128\\
Block2Conv2 &	Conv2D-128\\
Block2Pool &	Maxpooling2D\\
Block3Conv1 &	Conv2D-256\\
Block3Conv2 &	Conv2D-256\\
Block3Conv3 &	Conv2D-256\\
Block3Pool	& Maxpooling2D\\
Block4Conv1 &	Conv2D-512\\
Block4Conv2 &	Conv2D-512\\
Block4Conv3 &	Conv2D-512\\
Block4Pool	& Maxpooling2D\\
Block5Conv1 &	Conv2D-512\\
Block5Conv2 &	Conv2D-512\\
Block5Conv3 &	Conv2D-512\\
Block5Pool &	Maxpooling\\

\hline
\end{tabular}
\caption{The pre-trained VGG16 network model removed the top layer.(sec 2.2)}
\end{table}

\begin{table}[tbp]
\centering  
\begin{tabular}{ll}  
\hline
Layer & Type of layer \\ \hline  

Input (512$\times$1$\times$1 feature vector)\\

Flatten1 & Flatten-512\\
Dense1	& Dense-256\\
Dropout1 &	Dropout-256\\
Dense2 &	Dense-2\\
Soft-max\\

\hline
\end{tabular}
\caption{The binary classifier without CNN for feature extracting.(sec 2.3)}
\end{table}

\begin{table}[tbp]
\centering  
\begin{tabular}{ll}  
\hline
Layer & Type of layer \\ \hline  

Input (50$\times$50 RGB image)\\

Block1Conv1 & Conv2D-64\\
Block1Conv2	& Conv2D-64\\
BlockPool &	Maxpooling2D\\
Block2Conv1 &	Conv2D-64\\
Block2Pool &	Maxpooling2D\\
Dropout1 & Dropout\\
Flatten1 & Flatten-10816\\
Dense1 & Dense-128\\
Dropout2 & Dropout\\
Dense2 & Dense-2\\
Soft-max\\

\hline
\end{tabular}
\caption{The convolutional neural network constructed by ourselves.(sec 2.3)}
\end{table}

\begin{table}[tbp]
\centering  
\begin{tabular}{lllllll}  
\hline
ROC curve & AUC & Y index & Best T & ACC & AEN & SPE \\ \hline  

Class 0 & 0.976 & 0.886 & 0.615 & 0.943 & 0.959 & 0.927\\
Class 1 & 0.976 & 0.886 & 0.417 & 0.943 & 0.924 & 0.961\\

\hline
\end{tabular}
\caption{The performance of the binary classifier based on VGG16 model.(sec 3.2) Receiver operating characteristic curve (ROC curve), Area under ROC curve (AUC), Youden’s index (Y index), Best threshold (Best T), Accuracy (ACC), Sensitivity (SEN), Specificity (SPE) were determined.}
\end{table}

\begin{table}[tbp]
\centering  
\begin{tabular}{lll}  
\hline
Picture & Result of the paper & Result of experts \\ \hline  
a & [0, 0, 0.01, 0.97, 0, 0.02, 0] & c5 \\
b & [0.16, 0, 0.08, 0.47, 0.06, 0.22, 0] & c0, c4, c5 \\
c & [0.30, 0.06, 0.05, 0.38, 0.17, 0, 0.04] & c2, c4, c5\\
d &	[0.31, 0.05, 0.03, 0.11, 0.11, 0.19, 0.19] & c6\\
e & [0.55, 0.27, 0.03, 0.03, 0.01, 0.09, 0.01] & c1\\
f & [0, 0, 0.29, 0.63, 0, 0.05, 0.03] & c2\\
g & [0.07, 0.04, 0.04, 0.11, 0.64, 0.04, 0.04] & c4, c5\\

\hline
\end{tabular}
\caption{ Comparison between the overall description from our method and the description results from experts.(sec 3.3) Papule (c0), cyst (c1), blackhead (c2), normal skin (c3), pustule (c4), whitehead (c5), nodule (c6). The result of the paper is same with the histograms in Figure 8. The expert's diagnosis gives the main symptoms of facial vulgaris acne not including normal skin.}
\end{table}

\end{document}